# MAVOT: Memory-Augmented Video Object Tracking

Boyu Liu*  Yanzhao Wang*  Yu-Wing Tai  Chi-Keung Tang
HKUST  Tencent Youtu  HKUST
{bliuag,ywangdg}@connect.ust.hk  yuwingtai@tencent.com  cktang@cs.ust.hk

## Abstract

*We introduce a one-shot learning approach for video object tracking. The proposed algorithm requires seeing the object to be tracked only once, and employs an external memory to store and remember the evolving features of the foreground object as well as backgrounds over time during tracking. With the relevant memory retrieved and updated in each tracking, our tracking model is capable of maintaining long-term memory of the object, and thus can naturally deal with hard tracking scenarios including partial and total occlusion, motion changes and large scale and shape variations. In our experiments we use the ImageNet ILSVRC2015 [18] video detection dataset to train and use the VOT-2016 [13] benchmark to test and compare our Memory-Augmented Video Object Tracking (MAVOT) model. From the results, we conclude that given its one-shot property and simplicity in design, MAVOT is an attractive approach in visual tracking because it shows good performance on VOT-2016 benchmark and is among the top 5 performers in accuracy and robustness in occlusion, motion changes and empty target.*

## 1. Introduction

Video object tracking, to keep track of a moving object in a video sequence, is a key research area in computer vision with important applications such as public surveillance and transportation control. A tracking method should stably and accurately track an object of interest, which must also run in real time in time-critical applications.

State-of-the-art video object tracking uses online learning with convolutional neural networks (CNNs). The model is expected to track an object by tuning relevant parameters online during the training stage. Online trackers have shown good performance in tracking objects that have no dramatic changes in appearance over time.

Nevertheless, online training in object tracking have a number of limitations. First, the training data for online training methods are just the frames of the video itself. With such limited data on the object's appearance the complexity of the learned model will also be limited, in contrast to deep learning models which are trained using a huge amount of data. Second, in online training, the weights of the model are refined after introduction of new appearances of the object. Online back-propagation for weight updates is still time consuming and will compromise real-time performance. Third, online models do not have long term memory for recording past information, thus they will suffer from error accumulation and easily confused by recent distractions such as partial or total occlusion. Figure 1 shows a total occlusion example. Fourth, a number of online methods are category specific and they cannot track general objects. Lastly, detection-based tracking approaches often do not model background, because their discriminative classifiers are trained to detect foreground objects.

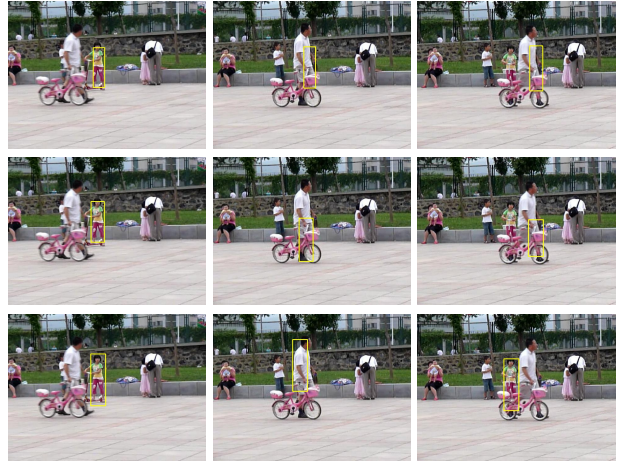

Figure 1. When an occluder (man) is moving slowly across the screen toward the tracking object (girl) as shown, top VOT-2016 [13] performers (e.g., CCOT [6] and SSAT [17]; see results section) start to track the occluder or lose the object at all. MAVOT uses one-shot learning and is robust to total occlusion, here continues tracking the girl when she reappears.

In this paper, we propose a new one-shot learning approach for visual object tracking: memory-augmented video object tracking (MAVOT). Given only an object bounding box in the first frame, MAVOT predicts the next in subsequent frames of the video. In [15], it is stated that our brain perceives moving objects to be farther along in their trajectory, or in other words, our vision *predicts* their location in current "frame". We believe our memory (visual cortex) plays a significant role in our one-shot learning ability to track objects even unseen before. Specifically, re-trieving from and updating the long-term memory allows us to effortlessly track a foreground object against its background, both of which may change rapidly over time, under partial or even total occlusion, severe motion change, and large scale variation. Inspired by our powerful memorization ability, the Google DeepMind among others proposed to attach to the deep neural network a memory module for longer term storage of key features [9, 20, 8], so that the model can perform learning tasks only given a limited training instances, or in other words, one-shot learning is achieved.

---
*Equal contribution.

Encouraged by the high promise, MAVOT is a deep neural network integrated with an external memory module for long term memory and tracking. MAVOT uses CNN to extract key features of an image region, communicates with its memory to score if it contains the target object and then updates its memory to remember its long-term appearance by adapting to its new appearance over time.

Our proposed MAVOT has following properties:

- **One-shot learning:** MAVOT can be trained even with one object instance.

- **Fast tracking:** MAVOT does not need any back-propagation to refine the network during tracking; MAVOT uses fully convolutional operations to boost speed.

- **Long-term memory:** MAVOT is empowered with ability to track in long sequence.

- **class-agnostic:** MAVOT is pre-trained using ILSVRC2015 video dataset, and can thus be deployed to track a large class of objects.

- **foreground and background:** MAVOT retrieves and memorizes the information of both foreground and background to improve tracking performance.

We evaluated MAVOT using VOT-2016 [13] as the benchmark. We have achieved rank 1 in accuracy on detecting motion changes and empty target, and rank 2 in robustness in occlusion and motion changes, while achieving a reasonable overall rank 16 among the 71 competitors in VOT 2016 challenge.

## 2. Related Work

Many video tracking models and outstanding algorithms have been proposed. Here we will review the approaches which have achieved state-of-the-art performance in object tracking and inspired MAVOT's design.

**Tracking-by-Detection and Online Learning.** The tracking problem can be formulated as an online learning problem [22, 23]: given an initial bounding box containing the target object, learn a classifier online and evaluate it at multiple locations in subsequent frames. Each new detection can be used to update the model. Despite the success of online methods, since the training data is the video itself, it inherently limits the richness of the model they can learn.

Early tracking-by-detection includes support vector tracking [1], random forest classifiers [19], and boosting [7, 2]. These classical methods have been made online for object tracking. Using a large number of image features, in [10] an online learning approach was proposed using structured output SVM and Gaussian kernels to directly predict the target's location. In [12], the online tracking task was decomposed into tracking (following the object from frame to frame), learning (estimating the detector's errors and updating the detector), and detection (localizing all observed appearance and correcting the tracker). Structural constraints were used to guide the sampling process of a boosting classifier. In [11], high-speed tracking was proposed using kernelized correlation filters. Using the circulant property of the data matrix can drastically reduce both storage and computation. In [4], rather than online learning, the authors propose a basic tracking algorithm to work with a fully-convolutional Siamese network trained end-to-end on the ILSVRC2015 dataset for object detection and tracking in video, which achieves real-time performance and state-of-the-art accuracy.

**Correlation Filter Based Tracking**. Object Tracking using adaptive correlation filters [5] is a frequently-used measure in online training of trackers. A region of interest is given in the first frame where features of this image region are extracted. Then a response map is generated. The response map indicates the object location in the image. After each prediction the output will be used to update the correlation filter. Many state-of-the-art tracking algorithms, including Fully-Convolutional Siamese Network [4] and Kernelized Correlation Filter tracking [11] have adopted this technique and achieved good results. However, the response map generated by correlation filters will be adversely affected by occlusion and abrupt appearance changes of the object.

**Tracking Algorithms using Memory-like Structure**. Existing video object tracking architectures attempt to store object appearance information and update models during online training. For example, Nam et al. [16] proposed an algorithm that models and propagates in convolutional neural networks in a tree structure called TCNN. This tracking system stores a CNN model in each branch of the tree. When estimating object states, it calculates a weighted average score of multiple CNNs , and the tree structure will be updated by adding new paths in the model. This algorithm achieves outstanding results in the VOT-2016 challenge and is one of the leading tracking algorithms. The main limitations with this model are: a tree structure of CNN is quite complicated, and the cost of training and updating model is very high.

**Neural Turing Machine and Memory-Augmented Neural Network**. The pertinent research in one-shot learning area includes Neural Turing Machine [9] and Memory Augmented Neural Network [20]. These models adopt the Von Neumann memory architecture and is differentiable so that gradient descent training can be applied on these structures. Recent progress shows that one-shot learning models can successfully perform tasks like learning priority sort algorithm, emulating an N-Gram model and classifying images. Moreover, one-shot learning can adapt to new conditions with only a few training examples provided. These properties are greatly desired in object tracking, but up to now there is no known representative work employing genuine one-shot learning in visual tracking tasks.

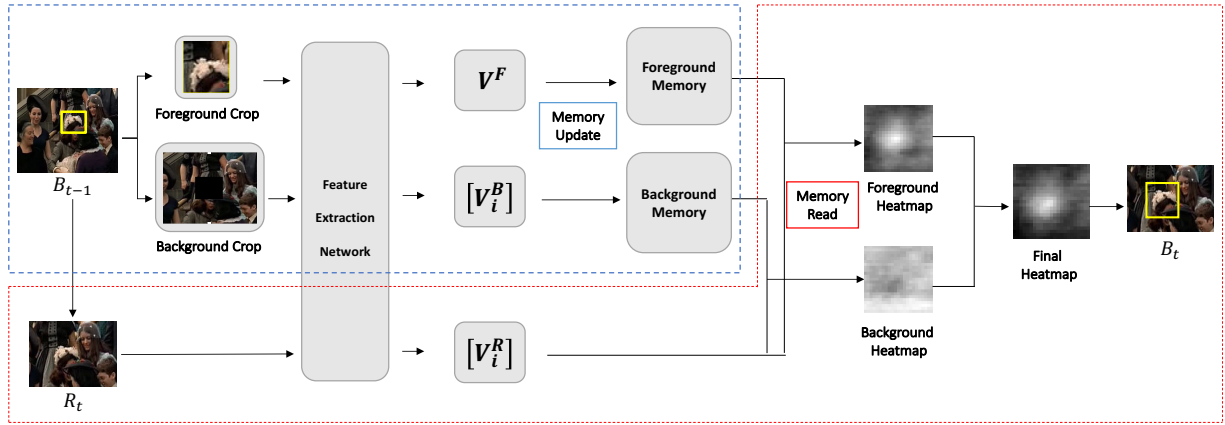

Figure 2. MAVOT consists of two main stages: object appearance update (blue) and bounding box prediction (red) for the current video frame. The first stage writes the memory and the second reads from the memory for tracking an object. The first stage includes Foreground/Background Sampling, Feature Extraction, and Memory Update. The second stage includes Region of Interest Generation, Feature Extraction, Memory Read, Heatmap Generation, and bounding box prediction. $B_{t-1}$: Bounding box region in frame $f_{t-1}$, $t = 1, \cdots$, shown here overlayed in $f_t$, $R_t$: Region of Interest in current frame $f_t$, $V_i^{\{R|F|B\}}$: Dimension reduced feature vector of patch $i$ in Region of Interest/Foreground/Background.

## 3. Overview

Figure 2 overviews the MAVOT pipeline consisting of two stages, which writes and reads from the memory respectively in each stage to update object appearance and predict the bounding box location in the current frame.

Note that 1) all MAVOT gets is the initial bounding box location $B_0$ in the start frame $f_0$ in our one-shot learning setting. Then MAVOT will iterate itself once it receives the visual information of subsequent frames, 2) MAVOT only performs forward operation without back propagation, thus is inherently faster than and fundamentally different from online training methods.

**Foreground and Background Sampling**. In frame $f_t$, MAVOT treats the bounding box region in the last frame $B_{t-1}$ as the foreground object to track. Then we sample patches around the bounding box as background patches.

**Region of Interest Generation**. For each subsequent frame, given the bounding box location $B_{t-1}$, MAVOT generates the region of interest (ROI) $R_t$ in the current frame. With the assumption that object displacement is not drastically large, searching within ROI is much faster than searching in the entire image. ROI normalization is done to make the network structure invariant to the input size.

**Feature Extraction Network**. Before any memory operations, each image patch is converted into its key feature vector. MAVOT uses a class-agnostic Feature Extraction Network, which is composed of three parts:

- **Mask**: Since the centroid of a given object is roughly at the center of its bounding box, MAVOT uses a Gaussian-distribution mask.
- **HALF-VGG**: MAVOT uses the first three blocks of VGG-16 [21] as the feature extractor, initialized with pre-defined VGG-16.
- **Dimension Reduction:** MAVOT deploys an offline trained dimension reduction layer to make the dimension of the feature vector consistent with the size of a memory blob in the memory module.

To further speed up MAVOT, we perform the feature extraction simultaneously on multiple patches in the ROI.

**Memory Modules and Operations.** MAVOT uses external memory modules to record key features of the target object as well as the surrounding background. To avoid irrelevant object features from interfering the accuracy of the heatmap and the final bounding box location, we use two memory modules, one for storing foreground and the other for background features. The two memories use the same memory read/write operations.

In the object appearance update stage 1, MAVOT will use the key features of the sampled patches to update the long term memory, in order to keep track of the evolving appearances of the foreground object as well as its surrounding background.

In the bounding box prediction stage 2, MAVOT will compare a given patch's feature vector derived from the current frame with the vector read from the memory, and encode their similarity into a probability for the patch to contain the target object. By performing this operation for all the patches within the ROI, a foreground and background heatmap will be respectively generated which will be used to determine the bounding box location.

We will describe the implementation details of each component in the following section.

# 4. Implementation

## 4.1. Foreground and Background Sampling

Given $B_{t-1}$, $t = 1, \cdots$, we perform foreground and background sampling. The foreground crop is given by the target object inside $B_{t-1}$. The background feature sampling is based on two assumptions: 1) The background patch should be close to the target object: since the search space of MAVOT is a region of interest generated around the target, the predicted object cannot be outside this range; 2) the background patch shares the feature that is similar to the target: MAVOT determines the target by measuring the similarity between image features and contents in memory. If MAVOT makes a wrong prediction (e.g. focus on the background scene or follow an object nearby), the feature of this wrong object should be similar with the correct object's feature. Thus, a background crop is generated around target object $B_{t-1}$, and detailed background feature selection methods will be elaborated in Section 4.4.4.

## 4.2. Region of Interest Generation

The ROI on the current frame $f_t$ is generated by zooming at the center of bounding box of the previous frame $B_{t-1}$ with a pre-defined ratio (now set at $160/360$ in this paper) to produce ROI $R_t$ which has the same height-width ratio as the given bounding box.

$B_0$, the only given data in our one-shot setting, is used to start MAVOT. MAVOT will iterate itself to predict following $B_t$. At times the ROI straddles across the image border. In this situation we will apply padding in the original image, and use the padded image to produce the ROI.

Normalization will be performed such that ROIs of all frames have the same size before feeding to the network. In our implementation, the shape of $B_t$ is $160 \times 160 \times 3$ and the shape of $R_t$ is $360 \times 360 \times 3$.

## 4.3. Feature Extraction Network

In MAVOT, Feature Extraction Network works in tandem with the memory module: in the object appearance update stage, relevant feature vectors are written to the memory; in the bounding box prediction stage, pertinent feature vectors are read from the memory to compare with current observations. Feature vectors are extracted from the target object or its surrounding background image patches.

To further boost time performance, multiple image patches are simultaneously processed. In the following, we first describe single-patch feature extraction to inspire our current implementation of the multiple-patch version.

### 4.3.1 Single-Patch Feature Extraction

The top of Figure 3 shows the single-patch version, which consists of three steps:

**Mask:** We assume the target object tends to be centered within a bounding box, so we apply a Gaussian mask at the center to give higher weight to the features near the center.

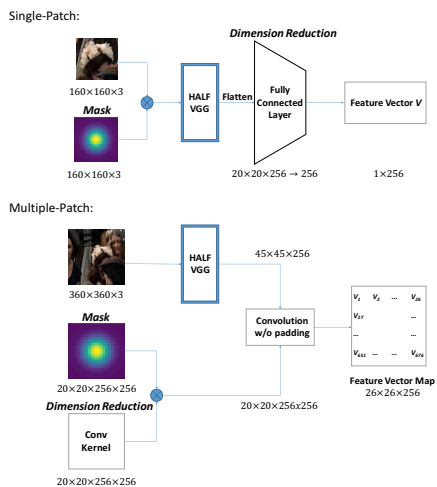

Figure 3. Feature Extraction Network. $V$ and $V_i$ respectively denote the key feature vector of a given patch. In the two versions, the module **Mask** and **Dimension Reduction** perform corresponding equivalent operations.

**Feature Extraction:** Rather than deploying the entire deep VGG16 [21] model we use HALF-VGG which contains only the first three blocks of VGG-16, so that the running time and memory usage can be significantly reduced. Also, the original VGG-16 has five pooling layers (which are not used in our HALF-VGG). After each pooling layer, the resolution will be halved, which will adversely affect the accuracy and tracking location.

The output of the Pool3 layer of VGG-16 is a feature map that is 1/8 the size of the original image, so the output will be $20 \times 20 \times 256$ with input $160 \times 160 \times 3$ for patches with the target object size.

**Dimension Reduction:** After feature extraction, the feature map of each patch is $20 \times 20 \times 256$, which is too large to write to our external memory module, as MAVOT will keep track of long-term appearance of the object (in other words, a lot of updated appearances) which can change significantly over time. Moreover, similarity metrics operating in such high dimension are not reliable. Thus we apply Fully-Connected Layer to achieve dimension reduction to make the final dimension the same as that of the memory blob, which is 256 here.

### 4.3.2 Multiple-Patch Feature Extraction

We might have forwarded each patch successively in the above single-patch feature extractor which would have been extremely inefficient. MAVOT uses a Multiple-Patch feature extractor as shown in next half of Figure 3, aiming at performing feature extraction for all patches in parallel. The following changes on the single-patch version are made:

**HALF-VGG:** As HALF-VGG is composed of only convolution and pooling operations which preserve spatial structure of all features, the only difference is that output of HALF-VGG will be $45 \times 45 \times 256$ with input $360 \times 360 \times 3$.

**Dimension Reduction Kernel:** Instead of using Fully-Connected Layer to perform dimension reduction, we reshape it into a convolution kernel. Applying convolution without padding is same as applying Fully-Connected Layer operation to all patches simultaneously.

**Mask:** Instead of masking a given patch, we apply the mask on the kernel.

Figure 3 illustrates that for the multiple-patch version, the result of dimension reduction is a $26 \times 26 \times 256$ tensor. Thus, the number of patches to examine is $26 \times 26 = 676$ patches where each patch is represented by a 256D feature vector after feature reduction.

### 4.4. Memory Modules and Heatmap Generation

The external memory module is the central part of MAVOT. Here we give detailed implementation, especially on addressing, reading, and writing mechanisms.

In MAVOT, there are two memory matrices respectively for the foreground (target object) and background. By introducing this dual memory structure, the tracking performance can be significantly improved. Specifically, using memory read/write operations to produce heatmaps on both memories to be detailed below, we can tell if MAVOT has started to detect a background object, and thus preventing it from tracking the background object when the target object is not clearly detected using the foreground memory.

Recall that given $B_0$ or $B_{t-1}$ in a subsequent frame, we feed the foreground/background crops to the Feature Extraction and Dimension Reduction Network, producing 256D vectors $V^F$ and $V^B$. These vectors will be written into their respective memories using the following mechanisms. In the following, unless otherwise stated the memory $M$, feature vectors $V$, and write/read from the a memory module are applicable to both foreground and background.

#### 4.4.1 Data Structure

In MAVOT, a memory module is represented by a matrix $M$, with the number of rows equal to the number of memory locations and the number of columns equal to the size of one memory blob ($128 \times 256$ in our implementation). Memory is updated after processing each frame, thus we use $M_t$ to represent memory for the current frame so far, where each memory blob contains a feature of the tracked object.

#### 4.4.2 Memory Write

Recall that given $B_0$ or a predicted bounding box in a subsequent frame, an object/background patch is represented by a 256D dimension-reduced feature vector $V$. In this section, we will first explain two mechanisms used to determine whether to write $V$ to the foreground/background memory. Then we will describe the erase and write operations for updating memory.

**Similarity-based write protection:** To decide if $V$ should be written, we check the maximum similarity of $S_V$ among all memory locations.

$$S_V = \max(M_t V) \quad (1)$$

If $S_V$ is larger than a threshold (we use 0.9), we will ignore this writing operation. This write protection mechanism can prevent the memory from being filled with almost the same content, so that more space will be left to write only sufficiently new appearance of the object.

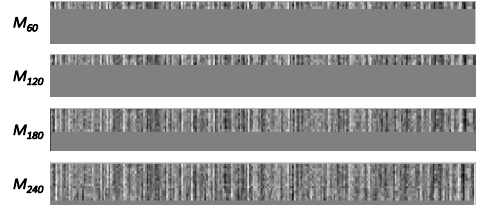

Figure 4. Memory blob visualization. From top to bottom, the foreground memory content at $t = 60, 120, 180, 240$. Feature vectors are written into memory as the tracking process continues while the long-term memory is being built.

**Least-recently-used addressing:** During tracking we maintain a usage weight $U_{t-1}$ to indicate each memory blob's usage frequency. Given $V$, when $S_V$ is smaller than a certain threshold (0.9 as stated), we will find the least recently used memory location as the write address to produce a one-hot write weight $W_V$.

The usage weight $U_{t-1}$ will be updated in three ways: 1) reset if a new memory location is written into, 2) add $S_V$, so that more similar memory blobs will be strengthened, and these blobs are used more frequently hence not likely to be overwritten, 3) decay over time:

$$U_t = (\gamma \times U_{t-1} + S_V) \otimes (1 - W_V) + W_V \otimes c \quad (2)$$

where $\gamma$ is a decay coefficient, $S_V$ is the similarity produced in Eq. (1), $\otimes$ is element-wise multiplication, $c$ is predefined constant for usage weight initialization (we use 1). This mechanism ensures to maintain the most useful memories when the memory is almost full.

Figure 4 visualizes a sample foreground memory module after successive updates. Very long sequences can be tracked despite significant object appearance changes and total occlusion using our memory update mechanism.

**Erase and Write:** With the one-hot writing weight $W_V$, the memory will erase the information of the location first:

$$M'_{t-1} = M_{t-1} \otimes ((1 - W_V)\text{ones}(256)) \quad (3)$$

where $\otimes$ means element-wise multiplication, and ones($\cdot$) denotes a vector with every element 1. Finally we write $V$ into the location:

$$M_t = M'_{t-1} + W_V V^T \quad (4)$$

#### 4.4.3 Memory Read

Given each patch $i$ in ROI $R_t$ in the current frame, we estimate its probability containing the foreground or background.

First, we will use dimension-reduced feature $V_i, i = 1 \cdots 676$, within $R_t$ to read from the memory. We adopt *content similarity-based* addressing for memory reading. Given feature vector $V_i$ and memory $M_t$, we calculate the read weight $r_i$ which is the exponential of cosine-similarity between $V_i$ and each memory blob in $M_t$:

$$r_i = \exp(M_t V_i) \quad (5)$$

Then, the memory readout given read vector $V_i$ is :

$$out_i = r_i^T M_t \quad (6)$$

### 4.4.4 Background Candidates Selection

Recall the two assumptions on background sampling: proximity and similarity to the foreground, which are both problematic issues even to state-of-the-art trackers. Therefore, we sample such background patches that are likely to cause prediction errors. Figure 5 shows a background crop from the previous frame, which is obtained by zooming at the center of the previous-frame bounding box by three times and masking out the region covered by the bounding box. We perform multiple feature extraction and dimension reduction as described in section 4.3 on this masked background crop, and produce a background candidate map using cosine-similarity. Here, *without* the foreground target object, high values in this candidate map indicates the background that shares high similarity with the key features in the foreground memory. The top 10 such patches are selected from the candidates and their respective features are written in the background memory, using memory write mechanism as described in section 4.4.

### 4.4.5 Heatmap Generation

For all of the $26 \times 26$ patch locations, we use the 256D feature vector as $V_i$ to read the memory, and obtain the read output $out_i$.

Here, we simply calculate the cosine-similarity of the two vectors and assign it as the foreground/background probability for the given patch with respect to the memory content. Collecting all $26 \times 26$ probabilities, we produce a heatmap that encodes the probability the patch containing the foreground/background.

Specifically, let $p_i^F$ and $p_i^B$ be respectively the foreground and background probability for patch $i, i = 1 \cdots 676$ in the ROI for the current frame. Then,

$$p_i^F = <V_i, out_i^F> \quad (7)$$
$$p_i^B = <V_i, out_i^B> \quad (8)$$

The final heatmap is computed by the subtracting them:

$$p_i = p_i^F - p_i^B \quad (9)$$

We subtract the background heatmap from the foreground heatmap because when MAVOT detects a background object (e.g. an occluder), patch $i$ where the occluder is present has a high $p_i^B$ and a low $p_i^F$. Hence the final heatmap $p_i$ should have a low value to make MAVOT avoid tracking the occluder. Note that there is no problem even at some locations $p_i^F < p_i^B$ because they will be discarded and only high values in the heatmap will be considered.

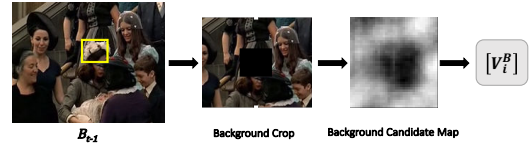

Figure 5. Background candidates generation and selection to write to the background memory.

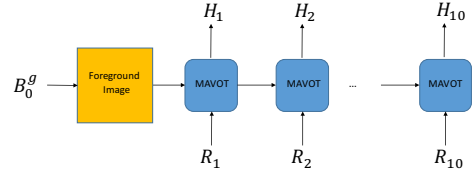

Figure 6. Training phase of MAVOT. Only the first ground truth is informed to the network, which comply with the One-Shot Learning setting. Notice that $H_t$ is just the set of all $p_i$'s in frame $f_t$.

Using the final heatmap which encodes the probability that each patch contains the target object, we simply pick the maximum among all $p_i$'s:

$$i^* = \arg\max_i p_i$$

Once the patch is obtained which has the highest $p_i$, the relevant foreground and background image crops will be sampled with their respective feature vectors extracted and dimension reduced. MAVOT then continues to track by updating the foreground and background memory as described in section 4.4.2. For tracking visualization, the coordinates of the corresponding patch will be warped back into the video frame space to produce the bounding box $B_t$ for the current frame.

### 4.5. Training

We use ImageNet ILSVRC2015 video detection dataset to train MAVOT. The Region of Interest $R_t$ is generated by the ground truth bounding box $B_{t-1}^g$ of $f_{t-1}$, and we train MOVOT on a sequence of length 10. Figure 6 shows the unrolled Recurrent Neural Network (RNN) with input sequence $R_{1,2\ldots10}$, output sequence of heatmaps of each frame $H_{1,2\ldots10}$, and a one-shot bounding box input $B_0$ on frame $f_0$.

With MAVOT's great simplicity in structure, the only part needs training offline is the Feature Extraction Network. For simplicity of training, we spare the background part and only write foreground features to memory, which also enables MAVOT's RNN to perform back-propagation with full differentiablity of every operation.

The loss of training is defined as the L2 difference between predicted heatmap $H_t$ and ground truth heatmap $H_t^g$,

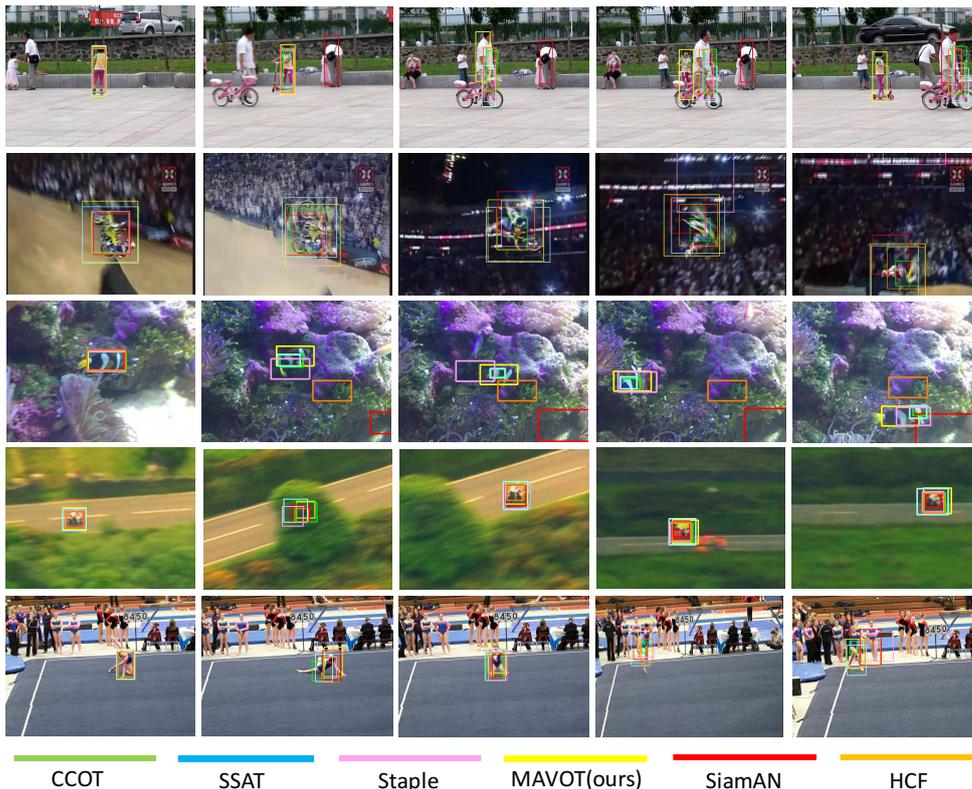

Figure 7. From top to bottom: *girl*, *motocross2*, *fish4*, *wiper*, *road*, *gymnastics4*. Full MAVOT tracking video demos are available in supplemental material.

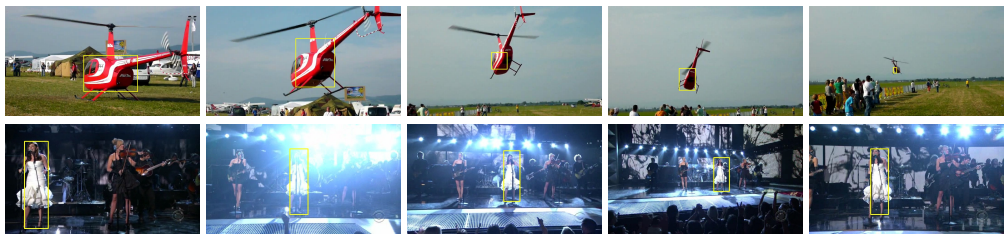

Figure 8. From top to bottom: *helicopter*, *singer*. Full MAVOT tracking video demos are available in supplemental material.

where $H_t$ is just the set of all $p_i$'s in $f_t$. $H_t^g$ is a 26 x 26 heatmap with Gaussian-distributed, center-aligned $B_t^G$ with standard deviation linear to the size of $B_t^G$.

## 5. Experiments

We employed the Visual Object Tracking (VOT [13]) toolkit for comparison. The VOT is a new but well-known challenge aiming at testing a competitor's performance on hard tracking examples and making statistical comparisons among different tracking algorithms. The dataset contains a variety of difficult tracking scenarios such as motion blur, occlusion, size and illumination changes. In our experiments, we use the latest published edition, VOT-2016 [13] challenge result, as our benchmark.

In VOT challenge, trackers are re-initialized when the Intersection of Union (IoU) between prediction and ground truth becomes zero. Under this setting, IoU is one evaluation criterion on accuracy (A). The other criterion is robustness (R), the number of re-initialization which corresponds to number of failures. The final ranking is given by the Expected Average Overlap (EAO) which is calculated by VOT algorithm. In addition to the challenge itself, VOT-2016 also has a setting where the toolkit will not intervene the testing tracker so that tracking can be performed to the very end. This additional setting allows us to visualize each tracker's tracking result and perform qualitative evaluation. As stated in the former sections, the evaluated MAVOT has the assumption that bounding box size will never change. However, we will also show MAVOT's potential power when the size of bounding boxes is allowed to change over time.

### 5.1. Qualitative Evaluation

We first present qualitative evaluation in this section. Figure 7 presents visual results of MAVOT in a number of difficult tracking scenarios. Here we compare our results with the state-of-art trackers: Continuous Convolution Operator Tracker (CCOT [6]), Scale-and-State Aware Tracker (SSAT [17]), Sum of Template And Pixel-wise

| | Overall Ranking | Accuracy A-Rank | | | | | | | Robustness R-Rank | | | | | | |
|---|---|---|---|---|---|---|---|---|---|---|---|---|---|---|---|
| | | overall | occ | camera | empty | illum | motion | size | overall | occ | camera | empty | illum | motion | size |
| CCOT | 0.3279[1] | 1.00 | 1.00 | 1.00 | 1.00 | 1.00 | 1.00 | 1.00 | 2.17 | 1.00 | 3.00 | 1.00 | 1.00 | 1.00 | 6.00 |
| SSAT | 0.3178[3] | 1.00 | 1.00 | 1.00 | 1.00 | 1.00 | 1.00 | 1.00 | 3.17 | 10.00 | 3.00 | 1.00 | 1.00 | 2.00 | 2.00 |
| Staple | 0.2924[5] | 1.00 | 1.00 | 1.00 | 1.00 | 1.00 | 1.00 | 1.00 | 11.17 | 22.00 | 7.00 | 6.00 | 12.00 | 14.00 | 6.00 |
| **MAVOT** | 0.2490[16] | 11.17 | 4.00 | 6.00 | 1.00 | 31.00 | 1.00 | 24.00 | 5.50 | 2.00 | 7.00 | 5.00 | 9.00 | 2.00 | 8.00 |
| SiamAN | 0.2326[22] | 1.00 | 1.00 | 1.00 | 1.00 | 1.00 | 1.00 | 1.00 | 11.50 | 14.00 | 17.00 | 6.00 | 8.00 | 22.00 | 2.00 |
| HCF | 0.2170[28] | 25.33 | 1.00 | 45.00 | 14.00 | 31.00 | 30.00 | 31.00 | 7.33 | 6.00 | 1.00 | 18.00 | 9.00 | 4.00 | 6.00 |

Table 1. VOT-2016 Challenge: MAVOT has achieved reasonable ranking (16th) given its simplicity which is arguably one of the first one-shot and deep-learning video object trackers. On the other hand, it has achieved rank 2 in robustness in occlusion, rank 1 in accuracy in empty and motion change, and top rankings in other categories. Denotations: occ-occlusion, camera–camera motion, illum–illumination change, motion–motion change, size–size change. Color: Red–1st rank, Blue–2nd rank, Green–3rd rank.

LEarners (Staple [3]), SiameseFC-AlexNet (SiamFC-A [4]) and Hierarchical Convolutional Features for Visual Tracking (HCF [14]). Video demos are in the supplemental material.

**girl**. MAVOT is robust against total occlusion. The *girl* is totally occluded by the passing man. While state-of-art trackers later track the occluder instead, MAVOT is able to find the original target and follow the target for a very long sequence (1500 frames).

**motocross2**. MAVOT is robust against large-scale shape changes and motion blurs. The *motorbike* moves and turns at a very high speed, causing image blur with its rapid change in shape. In spite of these difficulties, MAVOT is still able to keep tracking the motor bike to the end.

**fish4**. MAVOT is robust against large-scale shape changes and chaotic background in poor light. The swimming *fish* shares a very similar appearance with the background which causes tracking difficulties even for humans. Its rapid shape change under low light makes it a very hard tracking case. However, MAVOT can still track the fish at a relatively higher accuracy than the other state-of-art trackers.

**road**. MAVOT is robust against total occlusion in a very long sequence. The racking *bike* passes under several occluders (trees) in a long sequence (558 frames). Though the bike is totally occluded by trees or other facilities, MAVOT can resume tracking the target as soon as it reappears.

**gymnastics4**. MAVOT is robust against confusing background and large-scale shape changes. While performing, the gymnast exhibits quite dramatic shape changes, but MAVOT can still locate her with a high accuracy. The background contains several other similarly-looking gymnasts which poses confusion to other trackers where they will later track the background gymnasts. In contrast, MAVOT keeps tracking the target gymnast till the very end.

Figure 8 includes results of MAVOT which allows change in bounding box sizes.

**helicopter**. This sequence demonstrates large-scale size and shape variation of the target object, from spanning across the whole screen to a few pixels, when the helicopter is taking off from the ground and finally flying away in the sky.

**singer**. The *singer* was "occluded" by saturation due to the strong spotlight and the low dynamic range of the video camera, making her appear and disappear and reappear in the sequence, while the camera is being moved on dolly and the target is being zoom out.

### 5.2. Quantitative Evaluation

Table 1 tabulates the comparison between MAVOT and state-of-art trackers, both in overall rankings and rankings in specific categories. There were in total 70 trackers in VOT-2016 challenge, and MAVOT has achieved the 16th place. Since MAVOT has a fixed bounding box, it is expected to have an average performance in size changes. In the VOT dataset, we found that a large number of cases exhibit illumination changes occur together with size changes (e.g. *singer*), which explains our normal performance in this category. MAVOT performs very well in occlusion, empty, and motion changes comparably to the state-of-the-arts.

Using a single NVIDIA GeForce GTX1060 GPU (due to our limited equipment budget), MAVOT implemented with Python without specific optimization achieves **5fps**. With state-of-art trackers like CCOT [6] and TCNN [16] running at about 1fps on NVIDIA GeForce TITAN X, MAVOT is expected to run in real time on comparable devices.

### 6. Conclusion

We have presented MAVOT, a one-shot learning approach that augments external memory modules for maintaining long-term memory of the foreground as well as the background. Given only the bounding box in the first frame, the pretrained deep network only needs to see once the object to be tracked. MAVOT is different from online trackers because it does not need backpropagation among other advantages discussed in the paper. We performed extensive qualitative and quantitative experiments on MAVOT. While our current ranking in the VOT-2016 challenge may not be among the top ten, given the simplicity and its fundamental advantages, in contrast to the top VOT-2016 performers which were carefully engineered on conventional approaches and their improvements, we believe MAVOT has great potential. We hope this first paper will spawn interest and future work.